\documentclass{article}
\usepackage{ijcai19}

\usepackage{times}
\usepackage{soul}
\usepackage{url}
\usepackage[hidelinks]{hyperref}
\usepackage[utf8]{inputenc}
\usepackage[small]{caption}
\usepackage{graphicx}
\usepackage{amsmath}
\usepackage{mathrsfs}
\usepackage{booktabs}
\usepackage{algorithm}
\usepackage{algorithmic}
\usepackage{amssymb}
\usepackage{multirow}
\title{Leap-LSTM: Enhancing Long Short-Term Memory for Text Categorization}

\author{
Ting Huang\and
Gehui Shen\And
Zhi-Hong Deng\footnote{Corresponding Author}\\
\affiliations
Key Laboratory of Machine Perception (Ministry of Education),\\
School of Electronics Engineering and Computer Science, Peking University\\
\emails
\{ht1221, jueliangguke, zhdeng\}@pku.edu.cn
}

\begin{document}

\maketitle

\begin{abstract}
Recurrent Neural Networks (RNNs) are widely used in the field of natural language processing (NLP), ranging from text categorization to question answering and machine translation. However, RNNs generally read the whole text from beginning to end or vice versa sometimes, which makes it inefficient to process long texts. When reading a long document for a categorization task, such as topic categorization, large quantities of words are irrelevant and can be skipped. To this end, we propose  Leap-LSTM, an LSTM-enhanced model which dynamically leaps between words while reading texts. 
At each step, we utilize several feature encoders to extract messages from preceding texts, following texts and the current word, and then determine whether to skip the current word.
We evaluate Leap-LSTM on several text categorization tasks: sentiment analysis, news categorization, ontology classification and topic classification, with five benchmark data sets. The experimental results show that  our model   reads faster and predicts better than standard LSTM. Compared to previous  models which can also skip words, our model achieves better trade-offs between performance and efficiency.

\end{abstract}

\section{Introduction}
The last few years have seen much success of applying RNNs in the context of NLP, e.g. sentiment analysis~\cite{AdaHT-LSTM}, text categorization~\cite{DBLP:journals/corr/YogatamaDLB17}, document summarization~\cite{DBLP:conf/acl/SeeLM17}, machine translation~\cite{DBLP:journals/corr/BahdanauCB14}, dialogue system ~\cite{DBLP:journals/corr/SerbanSBCP15} and machine comprehension~\cite{DBLP:journals/corr/SeoKFH16}. A basic commonality of these models is that they always read all the available input text without considering whether all the parts of them are related to the task. For certain applications like machine translation,  it is a prerequisite to read the whole text. However, for text categorization tasks,  a large proportion of words are redundant and helpless for prediction.

As we know, training RNNs on long sequences is often challenged by vanishing gradients, inefficient inference and the problem in capturing long term dependencies. All three challenges are tightly related to the long computational graph resulting from their inherently sequential behavior of standard RNN.  Gate-based units as the Long Short-Term Memory (LSTM)~\cite{HochSchm97} and the Gated Recurrent Unit (GRU)~\cite{DBLP:conf/ssst/ChoMBB14} were proposed to address the problem of vanishing gradients and capturing long term dependencies. These two units are widely used as basic components in NLP  tasks because of their excellent performances.  However, they still suffer from slow inference while reading long texts.

Inspired by human speed reading mechanism, previous works~\cite{LSTM-jump,skip-rnn,skim-RNN} have proposed several RNN-based architectures to skip words/pixels/frames for processing long sequences.
When processing texts, their models  only consider the previous information and skip multiple words in one jump.
The biggest downside is that they are not aware of which words are skipped. It makes their skipping behavior reckless and risky.

In this paper, we  focus on skipping words for more efficient text reading on the task of text categorization. We present a novel modification, named Leap-LSTM, to the standard LSTM, enhancing it with the ability to dynamically leap between words.
``Leap" not only means that the model can leap over words, but also a leap on LSTM.
Previous models do not make full use of the information from the following texts and the current word, 
but we think they are important.
In our model, we fully consider the useful information at various aspects. More specifically, we design efficient feature encoders to extract messages from preceding texts, following texts and the current word for the decision at each step.

In the experiments, we show that our proposed model can perform skipping behavior with strictly controllable skip rate by adding a direct penalization term in the stage of training, which also means controllable and stable speed up on standard LSTM. Compared to previous works, our model tends to skip more unimportant words and perform better on predicting the category of the text. Moreover, we enhance standard LSTM with a novel schedule-training scheme to explore the reason why our model works well in some cases.  
In summary, our contribution is three-fold, which can be concluded as follows:
\begin{itemize}
    \item We present a novel modification to standard LSTM, which learns to fuse information from various levels and skip unimportant words if needed when reading texts.
    \item Experimental results demonstrate that Leap-LSTM can inference faster and predict better than LSTM. Compared to previous models which also skip words, our model skips more  unimportant words and achieves better trade-offs between performance and efficiency.
    \item We explore the underlying cause of our performance improvement over standard LSTM. According to the extensive experiments, we provide a new explanation of performance improvement, which has not been studied in previous works.
\end{itemize}

\section{Related Works}
\label{rw}
In this section, we introduce some previous works, which aims for efficient long sequence processing in the stage of training or practical applications.

Some works focus on adjusting the computation mode of standard RNN.
\cite{VCRNN} proposes Variable Computation RNN (VCRNN), which can update only a fraction of the hidden states based on the current hidden state and input.
\cite{CWRNN} and \cite{Phased-LSTM} design their models following the periodic patterns. ~\cite{CWRNN} presents Clockwise RNN (CWRNN) to partition the hidden layer into separate modules with different temporal granularity, and making computations only at its prescribed clock rate.
~\cite{Phased-LSTM} introduces the Phased LSTM model, which extends the LSTM unit by adding a new time gate controlled by a parametrized oscillation with a frequency range that produces updates of the memory cell only during a small percentage of the cycle.
However, these attempts were figured out that they  generally have not accelerated inference as dramatically
as hoped, due to the sequential nature of RNNs and the parallel computation capabilities of modern hardware ~\cite{skip-rnn}.

From another perspective, sevaral recent NLP applications have explored the idea of filtering irrelevant content by learning to  skip/skim words.
LSTM-Jump~\cite{LSTM-jump} predicts how many words should be neglected, accelerating the reading process of RNNs.
Skip RNN~\cite{skip-rnn} is quite similar to LSTM-Jump.
These two models skip multiple steps with one decision and they  jump only based on current hidden state of RNNs.
In other words, their models don't know what they skip.
However, our proposed Leap-LSTM leaps step by step  and multiple-step leap is not allowed. 

Most related to our work is Skim-RNN ~\cite{skim-RNN}, which predicts the current word as important or unimportant at each step. It uses a small RNN for unimportant words and a large RNN for important ones. So strictly speaking, Skim-RNN  does not skip words, only treats important and unimportant words differently.
As mentioned above, in accelerating inference, directly skipping is more effective than reducing the size of the matrices involved in the computations performed at each time step.

Early stopping behavior is also modeled to accelerate inference. LSTM-Jump~\cite{LSTM-jump} and  Skip RNN~\cite{skip-rnn} integrate the same early stopping scheme into their jumping mechanism, the reading will stop if {\tt 0} is sampled from the jumping softmax.
In the context of question answering,~\cite{DBLP:conf/kdd/ShenHGC17} dynamically determines whether to continue the comprehension process after digesting
intermediate results, or to terminate reading when it concludes that existing information is adequate to produce an answer. Our model does not use this technique.

\section{Methodology}
In this section, we describe the proposed Leap-LSTM. We first present the main architecture, then introduce the details of some components of the model. 
At the end of this section, we compare our approach with previous models which also skip words.

\begin{figure*}[htbp]	
\centering
\includegraphics[width=0.87\textwidth]{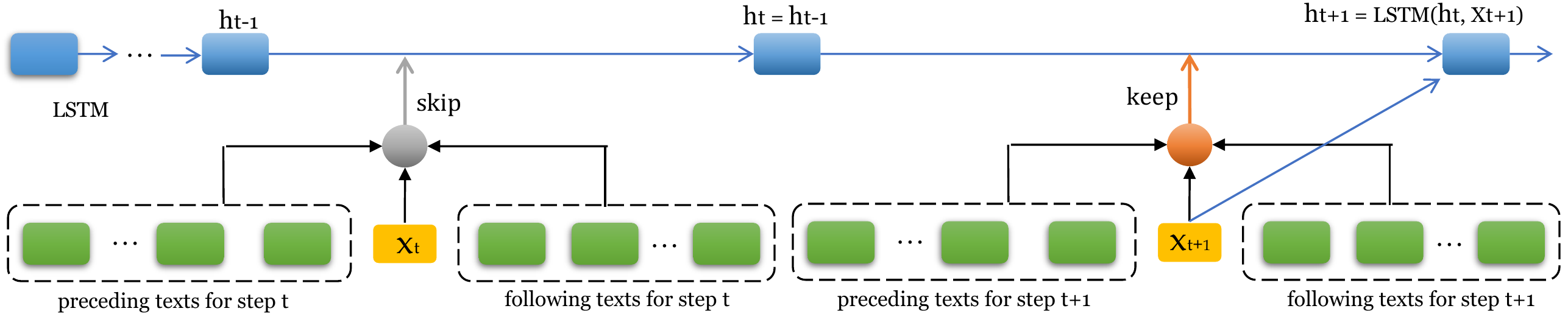}
\caption{An overview of Leap-LSTM. The small circle indicates the decision for skipping or keeping. In the example, Leap-RNN decides to skip at step $t$ and keep at the next step. So, the model directly copies $h_t$ from $h_{t-1}$, and conducts standard LSTM update functions  $\operatorname { LSTM } \left( \mathbf { h } _ { t } , \mathbf { x } _ { t+1 } \right)$ to obtain $h_{t+1}$.}
\label{model}
\end{figure*}
\subsection{Model Overview}
The main architecture of our model is shown in Figure \ref{model}. The model is based on the standard LSTM.
Given an input sequence denoted as $x_1, x_2, ..., x_T$ or $x_{1:T}$ with length $T$ (for simplification, we denote $x_t\in\mathbb{R}^d$ as the word embedding of the word at position $t$), our model aims to predict a single label $y$ for the entire sequence, such as the topic or the sentiment of a document.
In a sequential manner, the standard LSTM reads every word and applies update function to refresh the hidden state:
\begin{equation}
{\bf h}_t =  \operatorname{LSTM}({\bf h}_{t-1}, {\bf x}_t) \in \mathbb{R}^h,
\end{equation}
then the last hidden state is used to predict the desired task:
\begin{equation}
\label{equ2}
y \thicksim \operatorname{softmax}({\bf W}{\bf h}_T) \in \mathbb{R}^k,
\end{equation}
where ${\bf W}\in\mathbb{R}^{k\times h}$ is the weight matrix of the prediction layer and $k$ represents the number of classes for the task.
Leap-LSTM does not directly update the hidden state, but first compute a probability of skipping. At step $t$, we combine messages from preceding texts ($x_{1:t-1}$), following texts ($x_{t+1:T}$) and the current word ($x_t$). We use the word embedding $x_t$ as the message from the current word and we will discuss the choice of feature encoders for other two aspects later in Section \ref{fea_enc}. Currently, we simply use $f_{precede}(t)\in\mathbb{R}^p$ and $f_{follow}(t)\in\mathbb{R}^f$ to denote these two features at step $t$. Then we apply a two-layer multilayer perceptron (MLP) to compute the probability distribution  of skipping or keeping:
\begin{equation}
\label{equ3}
\setlength{\belowdisplayskip}{-5pt}
{\bf s}_t = \operatorname{RELU}({\bf W}_1[{\bf x}_t;{\bf f}_{precede}(t);{\bf f}_{follow}(t)] + {\bf b}_1),
\end{equation}
\begin{equation}
{\bf  \pi}_t = \operatorname{softmax}({\bf W}_2 {\bf s}_t+{\bf b}_2),
\end{equation} 
where \{${\bf W}_1\in\mathbb{R}^{s\times(d+p+f)}$, ${\bf W}_2\in\mathbb{R}^{2\times s}$, ${\bf b}_1\in\mathbb{R}^s$, ${\bf b}_2\in\mathbb{R}^2$\} are weights and bias of the MLP. $[;]$ denotes the vector concatenation.
${\bf s}_t\in\mathbb{R}^s$ is the hidden state of the MLP and ${\bf  \pi}_t\in\mathbb{R}^2$ represents the probability.
For the efficiency of inference, we need to make sure that the computation complexity of the MLP is substantially less than LSTM's updating function $O(h^2+hd)$. In our experiments, we set $s\ll h$ and $s \ll d$.

Obtaining the probability distribution ${\bf  \pi}_t$, two dimensions of it represent the probability of skipping and keeping respectively.
We sample a decision $d_t$ from ${\bf  \pi}_t$, $d_t=0$ means {\tt keep} and $1$ means {\tt skip}.
Formally, our unit updates the hidden state as:
\begin{equation}
{\bf h}_{t} =
\begin{cases}
\operatorname{LSTM}({\bf h}_{t-1}, {\bf x}_{t}) &\text{if $d_t = 0$}\\
{\bf h}_{t-1}  &\text{if $d_t = 1$,}
\end{cases}
\end{equation}
After processing the whole sequence, Equation (\ref{equ2}) is applied for predicting.

To control the skip rate of our model, we add a simple and straightforward penalty term to the final objective. It is proved to be very effective in the experiments. Formally, the loss function of Leap-LSTM is:
\begin{equation}
\mathcal { L } = \mathcal { L } _ {c} + \lambda (r_t - \mathit{r}) ^ {2},
\end{equation}
here $\mathcal { L } _ {c}$ is the loss from the classifier. $r$ denotes the total skip rate on the whole data set, while $r_t$ is our desired target rate. $\lambda$ serves as the weight of the penalty term. $\lambda$ and $r_t$ are hyperparameters to be set manually.

\subsection{Efficient Feature Encoders}
\label{fea_enc}
As the encoders for features used in Equation (\ref{equ3}), a desired characteristic of them is high computation efficiency.
If we use a complicated network as the encoder, our model just loses the advantage of fast inference, which violates the original intention of model design. 

For preceding texts, we reuse $h_{t-1}$, as $h_{t-1}$ encodes the information of all processed words.
We don't bring any extra computational cost for the message from preceding words.
For following texts, we partition it into two levels: local and global.
Take ${\bf x}_t$ in ${\bf x}_{1:T}$ for an example, we refer ${\bf x}_{t+1:t+m}$ as the local following text, and ${\bf x}_{t+1:T}$ as the global one.  

Convolutional neural networks (CNNs) have been used extensively in the context of NLP ~\cite{CNN,VDCNN} and we are impressed with their ability for extracting local patterns. 
The key  point is that CNNs have high computation parallelism, because they reuse the parameters (filter kernels) for each local regions.
Unlike RNNs, CNNs have no dependencies between different input regions.
We apply CNNs to encode local features, i.e. n-gram features of ${\bf x}_{t+1:t+m}$. We find in our experiments that CNNs process much faster than RNNs  with a similar amount of  parameters.

In order to extract all the following content, we employ a  reversed tiny LSTM with $h^\prime$-dimensional hidden state, where $h^\prime\ll h$. The reversed LSTM reads from the end of the sequence and generates an output at each step. 
We use $\operatorname{LSTM}_r(t^\prime)$ and $\operatorname{CNN}(t^\prime)$ to represent the output at step $t^\prime$  from reversed LSTM  and CNN respectively.
Note that here $\operatorname{LSTM}_r(t^\prime)$ encodes the features of ${\bf x}_{t^\prime:T}$, while $\operatorname{CNN}(t^\prime)$ encodes the features of ${\bf x}_{t^\prime:t^\prime+m}$. The following text features are obtained by
\begin{equation}
{\bf f}_{follow}(t) =
\begin{cases}
[ \operatorname{LSTM}_r(t+1); \operatorname{CNN}(t+1)] & \text{if $t < T$}\\
{\bf h}_{end} &\text{if $t = T$,}
\end{cases}
\end{equation}
where ${\bf h}_{end}\in\mathbb{R}^{f}$  is the representation of  features when the sequence reaches the end. So the desired information stored in it is the ending notification.
${\bf h}_{end}$ needs to be learned along with other model parameters during training.

\subsection{Relaxation of Discrete Variables}
Since we need to sample the decision $d$ (skip or keep) from categorical distribution ${\bf \pi}$, the model is difficult to train because the backpropagation algorithm cannot be applied to non-differentiable layers. 
We use gumbel-softmax distribution~\cite{gumbel-softmax,DBLP:journals/corr/MaddisonMT16} to approximate ${\bf \pi}$, which is also applied in  ~\cite{skim-RNN}.
Let $z$ be a categorical variable with class probabilities ${\bf \pi}_1, ..., {\bf \pi}_k$.
Gumbel-softmax distribution provides a simple and efficient way to draw samples $z$  from a categorical distribution with class probabilities  ${\bf \pi}$:
\begin{equation} 
z = \texttt{one\char`_hot} (\arg \max_i [g_i+log\left ({\bf \pi}_i \right)] ),
\end{equation}
where $g_1, ..., g_k$ are  i.i.d samples drawn from $Gumbel(0,1)$\footnote{Sampled as $g_i = -log(-log(u_i)), u_i\sim Uniform(0,1)$}.
We use the softmax function as a continuous  approximation to $arg\ max$, and generate  sample vectors ${\bf  y} \in \Delta ^ { k - 1 } $ ($(k-1)$-dimensional simplex) where
\begin{equation}
{\bf y} _ { i } = \frac { \exp \left( \left( \log \left( \pi _ { i } \right) + g _ { i } \right) / \tau \right) } { \sum _ { j = 1 } ^ { k } \exp \left( \left( \log \left( \pi _ { j } \right) + g _ { j } \right) / \tau \right) } \quad \text { for } i = 1 , \ldots , k,
\end{equation}
here $\tau$ is the softmax temperature. Finally, the update function of our unit can be represented as
\begin{equation}
{\bf h} _ { t } = [{\bf y} _ { t } ]_ 0 \cdot  \operatorname{LSTM}  \left({\bf h} _ { t -1 } , {\bf x} _ { t } \right) + [{\bf y} _ { t } ]_ 1 \cdot {\bf h} _ { t -1},
\end{equation}
then our model would be fully differentiable.

\subsection{Differences with Related Models}
\label{Differences with Related Models}
The biggest difference between LSTM-Jump, Skip RNN and our model  lies in the skipping behavior and processing of discrete variables. In LSTM-Jump and Skip RNN, the unit fails to consider the current word before jumping.
So their models skip multiple words at one step and then force the models to read regardless of what the next word is.
Intuitively, it could be a better choice to know all the contents before you decide to skip.
To this end, our model ``looks before you leap". We skip word by word and fuse the information from three aspects before skipping. 

To train the neural networks with discrete stochastic variables, we apply  gumbel-softmax approximation to make the whole model fully differentiable.
LSTM-Jump recasts it as a reinforcement learning problem and approximates the gradients with REINFORCE~\cite{REINFORCE}.
However, it is known to suffer from slow convergence and unstable training process.
Skip RNN applies the straight-through estimator~\cite{STE}, which approximates the step function by the identity when computing gradients during the backward pass.

Compared with Skim-RNN, our model skips words directly while  Skim-RNN uses a small RNN to process so-called unimportant words.

\section{Experiments}
We evaluate Leap-LSTM  in the field of text categorization. Especially, we choose five benchmark data sets, in which the text sequences are long. 
We compare our model with the standard LSTM and three competitor models: LSTM-Jump~\cite{LSTM-jump}, Skip RNN ~\cite{skip-rnn} and Skim-RNN~\cite{skim-RNN}.
To make the RNN units consistent, we use Skip LSTM  and Skim-LSTM to denote Skip RNN and Skim-RNN.

\subsection{Data}
We use five freely available large-scale data sets introduced by ~\cite{charCNN}, which cover several classification tasks (see Table \ref{datasets}). We refer the reader to ~\cite{charCNN} for more details on these data sets.
\subsection{Model Settings}
In all our experiments, we do not apply any data augmentation or preprocessing except lower-casing. 
We utilize the 300D GloVe 840B vectors ~\cite{GLOVE} as the pre-trained word embeddings. For words that do not appear in GloVe, we randomly initialize their word embeddings. Word embeddings are updated along with other parameters during the training process.


We use Adam ~\cite{Adam} to  optimize all trainable parameters with a initial learning rate $0.001$.
Dimensions \{$h, d, p, f, s, h^\prime$\} are set to \{$300, 300, 300, 200, 20, 20$\} respectively.
The sizes of CNN filters are \{[3, 300, 1, 60], [4, 300, 1, 60], [5, 300, 1, 60]\}. 
The temperature $\tau$ is always 0.1.
For $\lambda$ and $r_t$, the hyperparameters of the penalty term, different settings are applied, which depends on our desired skip rate. 
Throughout our experiments, we use a size of 32 for minibatches. 

\begin{table}[tbp]
\renewcommand\arraystretch{0.9}
\newcommand{\tabincell}[2]{\begin{tabular}{@{}#1@{}}#2\end{tabular}}
\begin{tabular}{@{}|c|c|c|c|c|c|@{}}
\hline
{\bf  Data set} & {\bf Task} & {\bf \#Classes} & {\bf \#Train} & {\bf \#Test} \\
\hline
AGNews	&  \tabincell{c}{News\\categorization} & 4 & 12k & 7.6k \\
\hline
DBPedia	& \tabincell{c}{Ontology\\classification} & 14& 560k & 70k \\
\hline
Yelp F.		& \tabincell{c}{Sentiment\\analysis} & 5 & 650k& 50k \\
\hline
Yelp P.		& \tabincell{c}{Sentiment\\analysis} & 2 & 560k & 38k \\
\hline
Yahoo		& \tabincell{c}{Topic\\classification} & 10 & 1400k & 60k \\
\hline
\end{tabular}
\caption{Statistics of five large-scale data sets. For each data set, we randomly select 10\% of the training set as the development set for hyperparameter selection and early stopping.}
\label{datasets}
\end{table}

\subsection{Experimental Results}
\begin{table*}[htbp]
\centering
\renewcommand\arraystretch{0.98}
\newcommand{\tabincell}[2]{\begin{tabular}{@{}#1@{}}#2\end{tabular}}
\setlength{\tabcolsep}{0.82mm}{ 
\begin{tabular}{@{}|c|c|ccc|ccc|ccc|ccc|ccc|@{}}
\hline
\multirow{2}*{Model} & \multirow{2}*{$r_t/\lambda$} & \multicolumn{3}{|c|}{AGNews} & \multicolumn{3}{|c|}{DBPedia} & \multicolumn{3}{|c|}{Yelp F.} & \multicolumn{3}{|c|}{Yelp P.} & \multicolumn{3}{|c|}{Yahoo}\\
\cline{3-17}
~ & ~ & acc & skip & speed & acc & skip & speed & acc & skip & speed & acc & skip & speed & acc & skip & speed\\
\hline
\multirow{4}*{Leap-LSTM}
& $0.0/0.1$ & 93.92 & 0.54 & 0.8x & {\bf 99.12} & $\approx 0$ & 0.8x & {\bf 66.50} & $\approx 0$ & 0.9x & {\bf 96.52} & $\approx 0$	 & 0.8x & 78.37 & $\approx 0$ & 0.9x \\
& $0.25/1.0$ & {\bf 93.93} & 24.93 & 1.1x & 99.10 & 27.94 & 1.2x & 65.91 & 22.71 & 1.1x & 96.20 & 23.57 & 1.1x & {\bf 78.40}& 26.89& 1.2x \\
& $0.6/1.0$ & 93.64 & 57.08 & 1.5x & 99.05 & 63.37 & 1.7x & 64.37 & 54.99 & 1.3x & 95.73 & 58.35 & 1.6x & 78.00 & 62.44 & 1.7x \\
& $0.9/1.0$ & 92.62 & 86.33 & 2.3x & 98.87 & 87.58 & 2.8x & 61.70 & 80.03 & 1.9x & 94.34 & 86.17 & 2.0x & 77.43 & 84.77 & 2.4x \\
\hline
\multicolumn{2}{|c|}{LSTM} & 93.23 & -  & 1.0x &  99.01 & -  & 1.0x &  65.93 & -  & 1.0x &  95.92 & -  & 1.0x &  77.92 & -  & 1.0x \\
\hline
\multicolumn{2}{|c|}{Skip LSTM}& 92.72 & 19.97 & 1.1x & 99.02 & 56.96 & 1.7x & 64.78 & 28.60 & 1.3x & 95.52 & 33.51 & 1.3x & 77.79 & 39.02 & 1.4x\\
\hline
\multicolumn{2}{|c|}{Skim-LSTM}& 93.48 & 49.66 & 1.3x & 98.61 & 73.10 & 2.1x & 65.08& 27.22 & 1.2x & 95.79 & 40.33 & 1.3x& 77.89 & 20.44 & 0.8x\\
\hline
\multicolumn{2}{|c|}{LSTM-Jump}& 89.30 & - & 1.1x &- & - & - & - & - & - & - & - & - & - & - & -\\
\hline

\end{tabular}
}
\caption{Test accuracy, overall skip rate and test time on five benchmark data sets. We apply four different \{$r_t, \lambda$\} settings for the different desired skip rate.   We implement  Skip LSTM and Skim-LSTM using their open-source codes. The results  reported in Skim-LSTM on AGNews are (93.60, 30.30, 1.0x). For LSTM-Jump, it uses REINFORCE to train the model and it performs poorly on AGNews. So we do not evaluate it on other data sets.}
\label{results}
\end{table*}

\begin{table}
\centering
\small
\renewcommand\arraystretch{1.0}
\newcommand{\tabincell}[2]{\begin{tabular}{@{}#1@{}}#2\end{tabular}}
\setlength{\tabcolsep}{0.9mm}{
\begin{tabular}{@{}|cccccc|@{}}
\hline
{\bf Model} & AGNews & DBPedia & Yelp F. & Yelp P. & Yahoo \\
\hline
LSTM               & 93.23	      & 99.01	   & 65.93   & 95.92    & 77.92    \\
\hline
\tabincell{c}{Leap-LSTM\\($r_t=0.0/\lambda=0.1$)}&{\bf  93.92} &{\bf 99.12} & {\bf 66.50} & 96.53 & 78.37 \\
\hline
\tabincell{c}{LSTM\\(schedule-training)} & 93.54 &  99.07 & 66.36 & {\bf 96.72} & {\bf 78.44} \\
 \hline
\end{tabular}
}
\caption{The accuracies of LSTM enhanced with schedule-training compared with LSTM and Leap-LSTM.}
\label{schedule-training} 
\end{table}

\subsubsection{Model Performances} 
Table \ref{results} displays the results of our model  and competitor models.
Each result is from the average of four parallel runs\footnote{Due to space limit, only the average results are shown in the table. See the appendix for complete experimental results. We provide a github link https://github.com/AnonymizedUser/appendix-for-leap-LSTM.}. 
Compared to LSTM, when skipping about 60\% or 90\% words, the decline in the accuracy is not significant. Our model even gets better accuracies on AGNews, DBPedia and Yahoo data with a speed up ranging from $1.5{\bf x}\sim1.7{\bf x}$.
When the desired skip rate is $0.0$ or $0.25$, the model improves LSTM across all tasks with an average relative error reduction of 8.0\% and 5.7\% respectively. 

Compared to other models which also skip words, our model  achieves better perfomances.
We can find that Leap-LSTM achieves better trade-offs between performance and efficiency.
For example, on AGNews data set, our model gets an accuracy of 93.64\% when skipping 57.08\% words and speeding up $1.5{\bf x}$.
In this case, our model reads faster and predict better than all competitor models. We attribute obtained improvements to our skipping behavior.
We will show it on several specific samples in later sections. 
The results also show that our penalty term works well. The model has the ability to control the overall test skip rate to lie in $[r_t-0.1, r_t+0.1]$, which means a stable and controllable speed up.
Compared with the regularization term used in other models, our method is more direct and controllable.

\begin{figure*}[htbp]	
\centering
\includegraphics[width=1.0\textwidth]{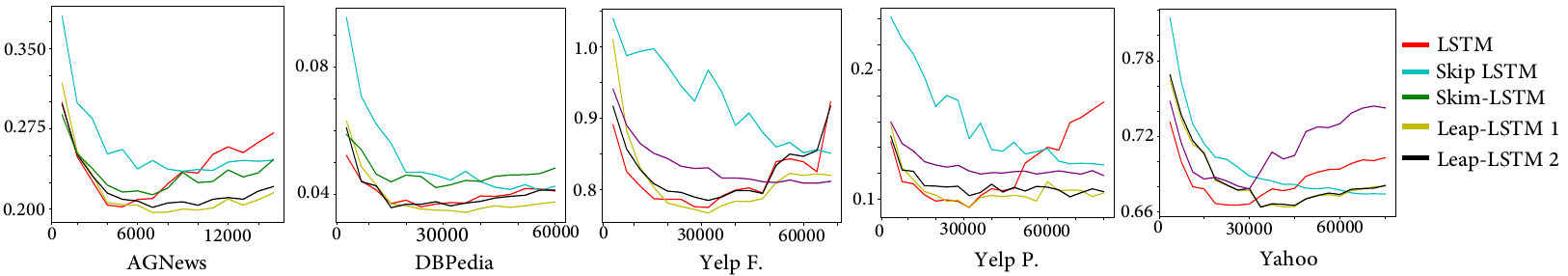}
\caption{The cross-entropy of the classifier on test set during training. Leap-LSTM $1\sim2$ corresponds to first two settings in table \ref{results} in turn.}
\label{loss_figure}
\end{figure*}

Figure \ref{loss_figure} displays the test cross-entropy of the classifier $\mathcal { L } _ {c}$ during training . We can find that LSTM converges faster, but  overfits early.
Our model (under the first two settings) consistently reaches lower loss on all data sets.
The curve of our model is much flatter than other models in the later stage of training.

\subsubsection{Schedule-Training}
In our experiments, one interesting thing is that Leap-LSTM obtains significant improvements over standard LSTM under the case \{$r_t=0.0, \lambda=0.1$\}. In this setting, almost no words are skipped when reading documents, but the classification accuracy improves. This phenomenon also happens in the experiments of  LSTM-Jump and Skip RNN. However, no reasonable explanation is given in their works.

In this paper, we hypothesize the accuracy improvement over LSTM could be due to the {\bf dynamic changes} of training samples. 
For a certain sample, the word sequence read by the LSTM unit is the same for each epoch when training LSTM. 
However, in the training stage of the models that can skip words, the unit reads different word sequences according to their skipping behavior.
Under the setting \{$r_t=0.0, \lambda=0.1$\}, we find that the overall skip rate on test set drops as the training goes on, and finally goes to zero. 
For the  LSTM cell used in our model, it sees  incomplete documents in the first few epoches.
To simulate the dynamic changes of training samples, we design a novel schedule-training scheme to train the standard  LSTM without changing the model.
Specifically, we randomly mask  words of each document with a probability $max(0, r_m-i*\beta)$ in the training set for epoch $i$. We set \{$r_m, \beta$\} to \{$0.45, 0.15$\}. The cell sees the whole documents from the third epoch on.
The schedule-training scheme is somewhat like dropout, and it may provide more different training samples to make the model generalized better.

The experimental results (see Table \ref{schedule-training}) demonstrate that LSTM with schedule-training consistently outperforms the standard LSTM on all five tasks, and gets close accuracies to Leap-LSTM. The improvement obtained by schedule-training scheme indicates that our hypothesis may make sense in the context of document classification. In addition, it also provides us a simple and promising way to improve RNN-based models without any modification to  original models.

\subsubsection{Ablation Tests}
We do ablation tests to explore what really matters for making an accurate decision.
Table \ref{ablation-test} shows the results of ablation tests under different settings on AGNews data set. From Leap-LSTM, we remove one component at a time and evaluate performance of partial models.

If removing \texttt{CNN} features or $\texttt{RNN}_r$ features seperately, the model still performs well when $r_t =0.0$ or $0.6$.
However, when  removing ${\bf f}_{follow}(t)$, the accuracy drops by  \{$0.23, 0.40, 0.25, 0.37$\}\% on four settings respectively, which indicates following text features are crucial for skipping words.
We can also find that preceding text features are helpful for skipping behavior, because of the large decline when removing ${\bf h}_{t-1}$. To our surprise, the word embedding of the current word ${\bf x}_t$ is not the most indispensable component, although it is also the important one.
Overall, all of these features make the full model perform more stably and achieve higher accuracies under all settings.
So, ``look before you leap" seems to be a good quality for neural networks.
\begin{table}[htbp]
\centering
\footnotesize
\renewcommand\arraystretch{1.0}
\newcommand{\tabincell}[2]{\begin{tabular}{@{}#1@{}}#2\end{tabular}}
\setlength{\tabcolsep}{0.6mm}{ 
\begin{tabular}{@{}|ccccc|@{}}
\hline
\bf 
\tabincell{c}{ Ablation\\Setting}&\tabincell{c}{ acc\\$(r_t=0.0)$}& \tabincell{c}{ acc\\$(r_t=0.25)$} &\tabincell{c}{ acc\\$(r_t=0.6)$}& \tabincell{c}{ acc\\$(r_t=0.9)$}\\
\hline
Full Model&					 		93.92& 		93.93&		93.64&		92.62\\
w/o \texttt{CNN} features&			 -0.06& 	-0.39& 		-0.08& 		-0.21\\
w/o $\texttt{RNN}_r$ features&		 -0.08& 	-0.46& 		-0.04& 		-0.09\\
w/o ${\bf f}_{follow}(t)$&			 -0.23&		-0.40&		-0.25&		-0.37\\
w/o  ${\bf h}_{t-1}$& 						-0.42&		-0.26& 		-0.03& 		-0.37\\
w/o ${\bf x}_t$& 							-0.30& 		-0.11& 		-0.01& 		-0.67\\
\hline
\end{tabular}
}
\caption{Ablation test for features used to predict a decision on AGNews data set, removing
each component separately. w/o ${\bf f}_{follow}(t)$ denotes the setting in which both \texttt{CNN} features and $\texttt{RNN}_r$ features are removed.} 
\label{ablation-test} 
\end{table}

\begin{figure*}[htbp]	
\centering
\includegraphics[width=0.99\textwidth]{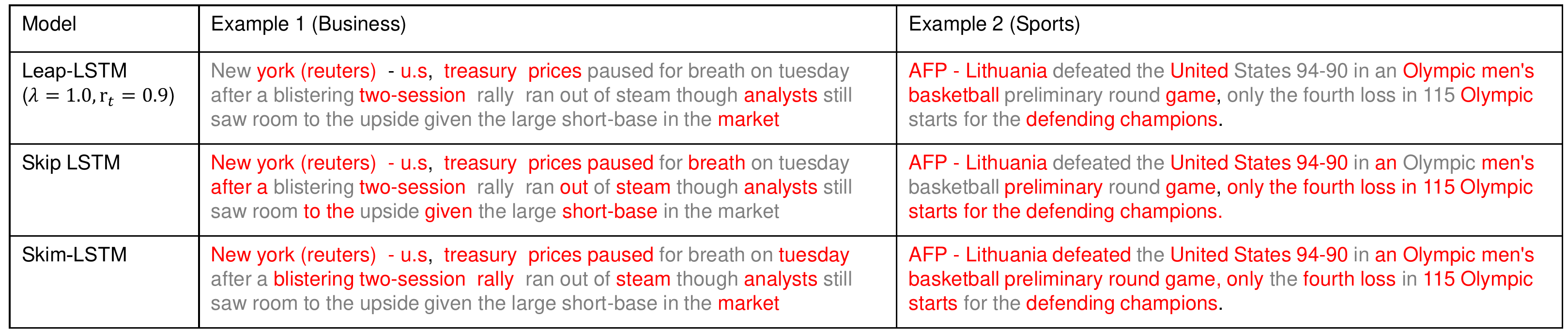}
\caption{Two examples from AGNews data set with Leap-LSTM ($ r_t=0.9, \lambda=1.0 $), Skip LSTM (default setting) and Skim-LSTM (default setting). They are from the topic Business and Sports. Words with grey color are skipped by the model and red ones are kept.}
\label{cases}
\end{figure*}

\subsubsection{Skipping Analysis}
In this part, we make a specific analysis of the skipping behavior of our model (under the setting \{$ r_t=0.9, \lambda=1.0 $\}). We display it from two  perspectives as follows.

\paragraph{Top-5 keep-rate words.} We count top-5 words in test keep rate\footnote{Computed as: times(kept) /   times(appear) on the whole test set.} for each class on DBPedia and display 6 classes of them (see table \ref{top-5}). We can find that the words most easily retained by our model are informative for reflecting the topic of the document.
For example, ``fc" (football club), ``club", ``playing", ``Olympics" are all widely-used words in the field of {\bf Athlete}. For the topic {\bf Film}, we can construct such a complete sentence with its top-5 keep-rate words: ``A movie {\it star} won the {\it award} at the {\it festival} with the {\it role} in this  {\it film}."
In addition, the keep rate of these words can even reach 100\%, which is not shown in the table.
These results demonstrate that our model has the ability to identify relevant words to the topic of the whole document, and then retain them. 
That is why our model has no performance degradation or even some improvement compared to the standard LSTM when skipping a large number of words.
Large quantities of irrelevant information are skipped through skipping behavior, making the model easier to infer the category of the document.

\paragraph{Case study. }
Figure \ref{cases} displays two examples, which are randomly selected from AGNews data set.
In example 1, Leap-LSTM retains all {\bf Business}-related terms in this piece of news: ``treasury prices", ``two-session", ``analysts", ``market".
As a result, our model can clearly classify this article into correct topic  with only less than 25\% of the words retained.
As for Skip LSTM and Skim-LSTM, they keep more words.
Obviously, many of them are helpless for predicting the topic, such as stop words (``a" and ``the") and prepositions (``after" and ``to").
In example 2, important words (phrases) like ``Olympic men's basketball", ``game", ``champions" are retained by Leap-LSTM. They are crucial for the model to predict this article as a {\bf Sports} news.
Our model skips most of the unimportant words.
Skip LSTM skips ``Olympic" and ``basketball", while Skim-LSTM skips only prepositions and three ``the"s.
In the term of skipping behavior,
Skip LSTM is not sure which types of words should be retained. It fails to identify relevant words and irrelevant words.
Skim-LSTM can retain important words as well as many unimportant ones.
Our model skips more words and more accurately than them.
We attribute it to the full consideration of features from various aspects, indicating that ``look before you leap" does help for more accurate skipping. 

\begin{table}[tbp]
\centering
\small
\renewcommand\arraystretch{1.0}
\newcommand{\tabincell}[2]{\begin{tabular}{@{}#1@{}}#2\end{tabular}}
\setlength{\tabcolsep}{0.95mm}{ 
\begin{tabular}{@{}|c|c|c|c|c|c|c|@{}}
\hline
Athlete & Building & Animal   & Album & Film &  WrittenWork \\
\hline
national	 &  register		& mollusk			& rock		& festival	& science \\
fc    		 	&  city			& wingspan			& records	& award	& fiction \\
club		& st 			& moist					& single	& stars		& comic \\
playing	& places		& noctuidae				& tracks	& role		& world \\
Olympics	& road			& forests				& band		& films		& edition \\
\hline
\end{tabular}
}
\caption{Top-5 highest keep-rate words of each class on DBPedia data set. We only display 6 classes of all 14 because of the page limit. Please see the full table in the appendix.}
\label{top-5}. 
\end{table}

\section{Conclusions}
In this paper, we present Leap-LSTM, an LSTM-enhanced model which can perform skip behavior with strictly controllable skip rate.
In the model, we combine messages from three aspects for skipping at each step.
Experimental results demonstrate that in the field of text categorization, our model predicts better previous models and the standard LSTM by skipping more accurately.
We conduct skipping analysis to explore its tendency for skipping words by some specific examples.  
Moreover, we design a novel schedule-training scheme to  train LSTM, and get close test accuracies to our model.
The improvement obtained by schedule-training indicates that our performance improvement over standard LSTM may due to the dynamic changes of training samples in addition to its ability to skip irrelevant words.
Our model is simple and flexible and it would be promising to integrate it into  sophisticated structures to achieve
even better performance in the future.

\section*{Acknowledgements}
This work is partially supported by the National High Technology Research and Development Program of China (Grant No. 2015AA015403).
We would also like to thank the anonymous reviewers for their helpful comments.

\bibliographystyle{named}

\end{document}